\documentclass[preprint,12pt]{elsarticle}
\usepackage{epstopdf}
\usepackage{graphicx}
\usepackage{multirow}
\usepackage{algorithm}
\usepackage{algpseudocode}
\usepackage{algorithmicx}
\journal{}
\begin{document}
\begin{frontmatter}
\title{Predicting Climate Variability over the Indian Region Using Data Mining Strategies}

\author[focal]{M.~Naresh Kumar\corref{cor1}}
 \ead{nareshkumar\_m@nrsc.gov.in}
 \cortext[cor1]{Principal Corresponding Author}
 \cortext[cor1]{Tel.: +91 40 2388 4388; Fax.: +91 40 2388 4437}
 \address[focal]{Software and Database Systems Group, National Remote Sensing Center (ISRO),  Hyderabad, Telangana,  500 037, India}

\begin{abstract}
In this paper an approach based on expectation maximization (EM) clustering to find the climate regions and a support vector machine to build a predictive model for each of these regions is proposed. To minimize the biases in the estimations a ten cross fold validation is adopted both for obtaining clusters and building the predictive models. The EM clustering could identify all the zones as per the Koppen classification over Indian region. The proposed strategy when employed for predicting temperature has resulted in an RMSE of $1.19$ in the Montane climate region and $0.89$ in the Humid Sub Tropical region as compared to $2.9$ and $0.95$ respectively predicted using k-means and linear regression method.
\end{abstract}
\begin{keyword}
support vector machine, expectation maximization, k-means, regression, climate regions, climate change, Koppen  classification
\end{keyword}

\end{frontmatter}

\section{Introduction}
Regionalization techniques are found to be effective in improving the prediction accuracies of the climate models. Building regional models and predicting the climate variability require processing and extraction of information from large volumes of high dimensional data sets. Data mining methods such as k-means (KM) clustering and statistical methods such as linear regression (LR) are popular techniques commonly employed for grouping the data into regions of similar climate and build a model to predict the climate variables for subsequent years. The k-means method requires specifying initial k clusters centers which is generally not known a priori. Also, the procedure is sensitive to the selection of the initial cluster centers. Moreover, a linear regression model may not capture the non-linear relationships among the climate variables. 

The EM finds clusters by finding a appropriate fit for the given data set with a mixture of Gaussians. Each of the Gaussians is associated with a mean and a covariance matrix. The prior probability for each Gaussian is computed as a total fraction of points in the cluster defined by that Gaussian. Based on the iterative approach in updating values for means and variances the optimal solution is reached.

In this paper an approach based on expectation maximization (EM) clustering to find the climate regions and a support vector machine to build a predictive model for each of these regions is proposed. To minimize the biases in the estimations a ten cross fold validation is adopted both for obtaining clusters and building the predictive models.

The following are the main objectives of the present work
\begin{enumerate}
\item Understand the process of climate change over Indian region through development of information extraction techniques that can effectively predict the climate variability
\item Develop a methodology for processing the long term gridded climate data and obtaining climate regions using  expectation maximization clustering

\item Prepare the maps of the climate regions identified by expectation maximization clustering and compare it with standard climate zones as per Köppen classification over Indian region

\item Evolve a procedure to subset the long term climate dataset into regional data sets 
\item Develop methods to extract the train data set for building the support vector regression classifier based on the number of years to predict
 \item Obtain the validation data set for each of the grid locations and compute the root mean squared error. 
\item Compare the performance of the proposed methodology with k-means and linear regression procedure
\end{enumerate}
This paper is organized as follows. In Section~\ref{meth} the proposed methodology of predicting climate variables is presented. The experiments and results are discussed in Section~\ref{expresults}. Conclusions and discussion are deferred to Section~\ref{concl}.

\section{Methodology}
\label{meth}
In the proposed methodology the climate dataset is first regionalized by applying Expectation maximization clustering using the long term averages of the climate variables. Further, a predictive model is developed using support vector machine SVM regression kernel. A ten cross fold validation is employed to obtain a robutst estimates of the root mean square error (RMSE).
\begin{figure}[H]
 \centering
  \includegraphics[width=3.0in]{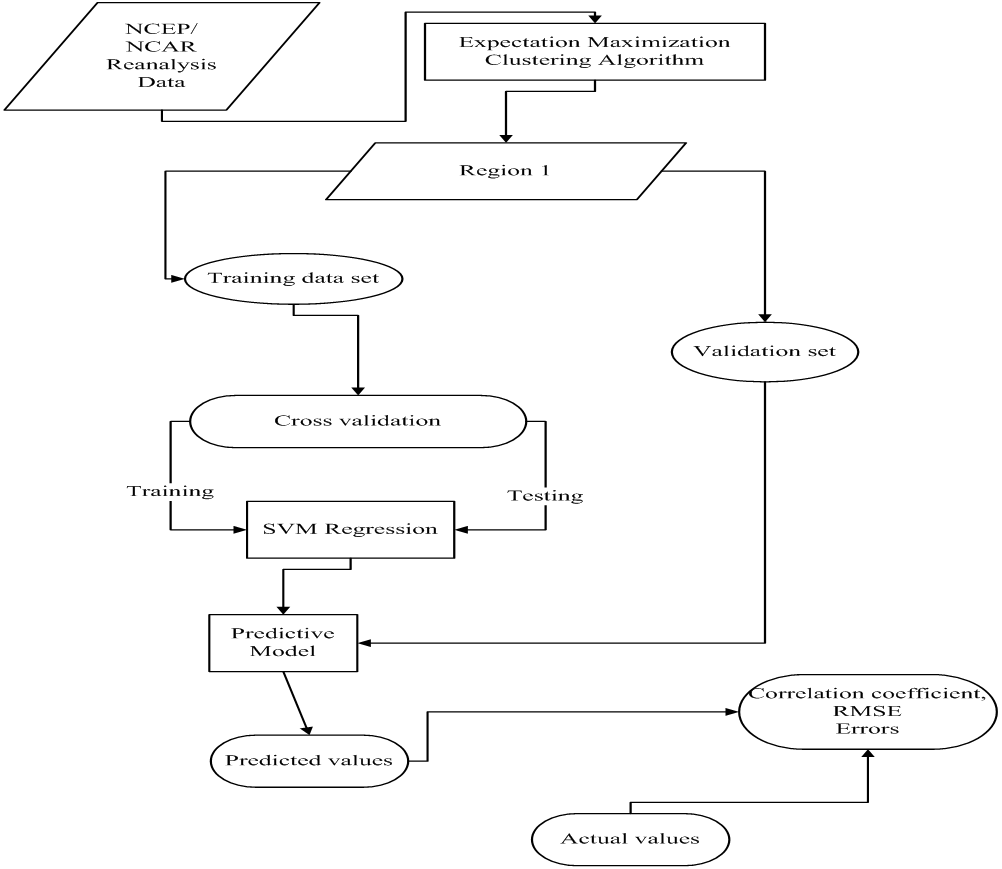}\\
  \caption{A flow chart depicting the procedure for building a predictive model}\label{fig1}
\end{figure}

The procedure employed in developing a predictive model is shown in Figure~\ref{fig1}.

The algorithm~\ref{alg1} describes the steps implemented in the present paper for obtaining a model for predicting climate variables. 
\begin{algorithm}
\caption{Procedure for Predicting Climate Variables}
\label{alg1}
\begin{algorithmic}
\Require
\begin{enumerate}
       \item  Climate data set
        \item Clustering method
        \item Number of years to predict (p)
         \item Variable to be predicted

\end{enumerate}
\Ensure
\begin{enumerate}
            \item Correlation coefficient
\item Root mean squared error

\end{enumerate}
   \textbf{Algorithm}
   \begin{enumerate}
      \item  Extract long term mean of climate variables for each $2.5^{\circ}$ x$2.5^{\circ}$ grid over Indian region
 \item Apply clustering method to obtain regions $R_1$, $R_2$,\ldots, $R_n$
\item Build the Model for the variable to be predicted

\begin{enumerate}
\item For each region in $R_1$, $R_2$,\ldots, $R_n$
\begin{enumerate}
\item obtain mean of the climate variables for all the grid points in the cluster for $j$-$p$ years where $j$ denotes total number of years and $p$ denotes number of years for which prediction is required
\item build a support vector machine regression model using a ten cross fold validation procedure
\end{enumerate}
\end{enumerate}
\item Test the model built in Step 3 

\begin{enumerate}
\item For each cluster in $R_1$, $R_2$,\ldots, $R_n$
\begin{enumerate}
     \item For each grid point in the cluster 
		\begin{enumerate}
     	    \item apply the corresponding model to predict precipitation and temperature for years $1$,\ldots, $p$
 \item compute the $RMSE$ using the predicted values and the actual values of the climate variables
 \end{enumerate}
		
   \end{enumerate}
\end{enumerate}
	\item RETURN $RMSE$.
    \item END.
		\end{enumerate}
\end{algorithmic}
\label{alg1}
\end{algorithm}

\section{Experiments and Results}
\label{expresults}
NCEP/NCAR reanalysis data for $65$ years from $1948$ to $2012$ having the climate variables Atmospheric Pressure, Relative Humidity, Precipitable Water, Zonal Wind, Meridional Wind, Precipitation, Air Temperature is used in the analysis

The application of the EM clustering on the dataset has resulted in $7$ climate regions. As per Koppen Classification only there are six regions. The present algorithm~\ref{alg1} has brought out a new region consisting of Uttaranchal , Sikkim and Arunachal Pradesh out of the existing Montane climate region. This we attribute it to the climate change and further investigations are required to ascertain these findings.

\begin{table}[!t]\scriptsize
\caption{EM Cluster Centriods of different Climate Zones}\label{tab1}
\begin{tabular}{p{0.9in}||p{0.7in}||p{0.5in}||p{0.5in}||p{0.5in}||p{0.5in}||p{0.5in}||p{0.5in}}
\hline 	  	 	 	 	 

\bfseries{Climate Variable}&\bfseries{Montanenew}&\bfseries{Semi Arid}&\bfseries{Tropical Wet and Dry}&\bfseries{Arid}&\bfseries{Montane}&\bfseries{Tropical Wet}&\bfseries{Humid Sub tropical}\\
\hline \hline
Air Temperature &	12.44	&27.04	&25.79&	25.81&	-2.54&	26.83&	24.8\\ \hline
Precipitable water 	&18.86	&41.32	&29.19	&22.02&	6.31&	38.01&	37.61\\ \hline
Precipitation &	4.8	&3.1	&2.57	&0.67&	2.36&	3.03	&6.4\\ \hline
Relative Humidity &	81.96	&76.73&	53.05&	35.19	&78.81	&75.51	&74.5\\ \hline
Sea Level Pressure &	1011.07	&1008.87	&1008.08	&1007.8&	1015.22&1009.76&	1009.43\\ \hline
Zonal winds 	&0.69	&0.94	&0.57	&1.1	&2.99	&2.67	&0.69\\ \hline
Meridional winds &	1.02	&1.63	&-0.36	&0.88	&1.82&	-1.01&	0.54\\ \hline

\end{tabular}
\end{table}

The cluster centroids for the seven regions are shown in Table \ref{tab1}. The air temperature in the montanenew regions is very high when compared to Montane region the reasons for under investigation.

\begin{figure}[H]
 \centering
  \includegraphics[width=3.0in]{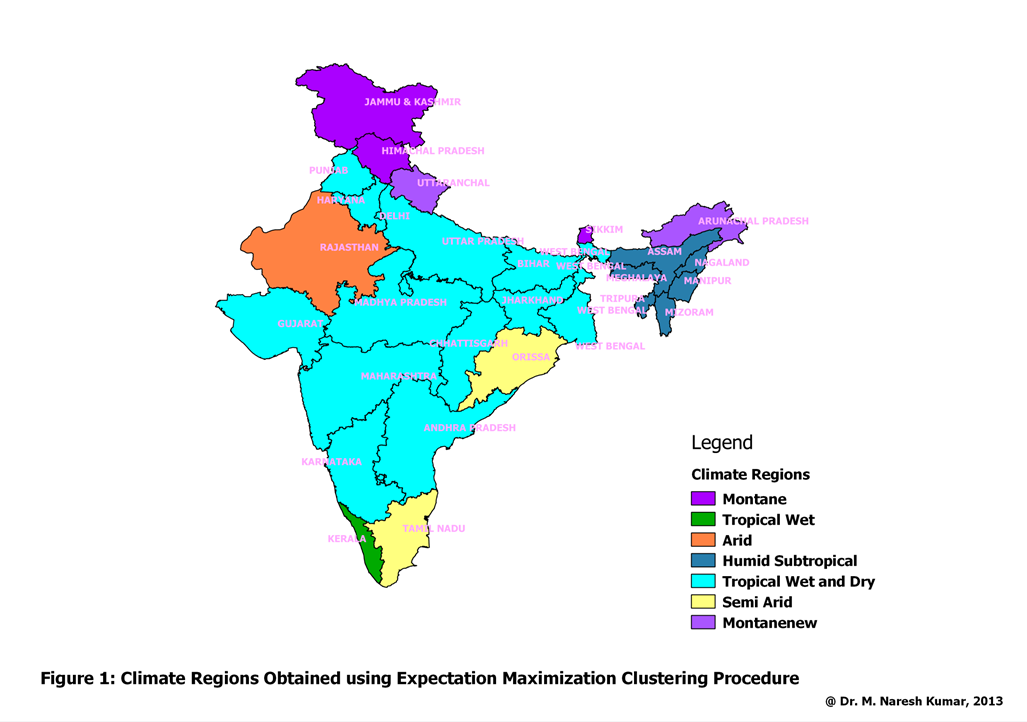}\\
  \caption{Climate Regions Obtained using Expectation Maximization Clustering Procedure}\label{fig2}
\end{figure}

The spatial extents of the climate regions obtained from proposed algorithm~\ref{alg1} is shown in Figure~\ref{fig2}.

\begin{table}[!t]\scriptsize
\caption{RMSE error for different climate zones}\label{tab2}
\begin{tabular}{p{0.9in}||p{0.7in}||p{0.5in}}
\hline 	  	 	 	 	 
 
\bfseries{Region}&\bfseries{EM+SVM}&\bfseries{KM+LR}\\
\hline \hline
Montanenew &	1.19	&2.9\\ \hline
Semi Arid 	&0.97&	0.88\\ \hline
Tropical Wet and Dry &	3.42&	2.94\\ \hline
Arid 	&0.68	&0.74\\ \hline
Montane 	&1.93&	1.69\\ \hline
Tropical Wet 	&0.55	&0.48\\ \hline
Humid Sub Tropical 	&0.8	&0.95\\ \hline

\end{tabular}
\end{table}
The RMSE errors in predicting the temperature for the year $2012$ is given in Table~\ref{tab2}.

\begin{figure}[H]
 \centering
  \includegraphics[width=3.0in]{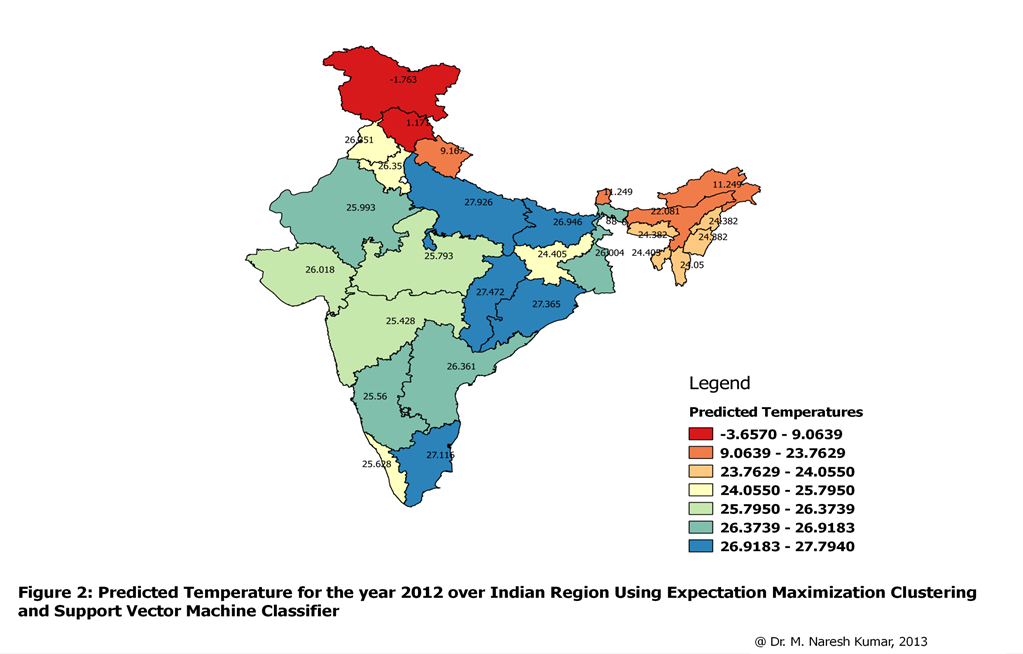}\\
  \caption{Predicted temperature for the year $2012$ over Indian region obtained using Expectation Maximization and SVM Regression Procedure}\label{fig3}
\end{figure}

\begin{figure}[H]
 \centering
  \includegraphics[width=3.0in]{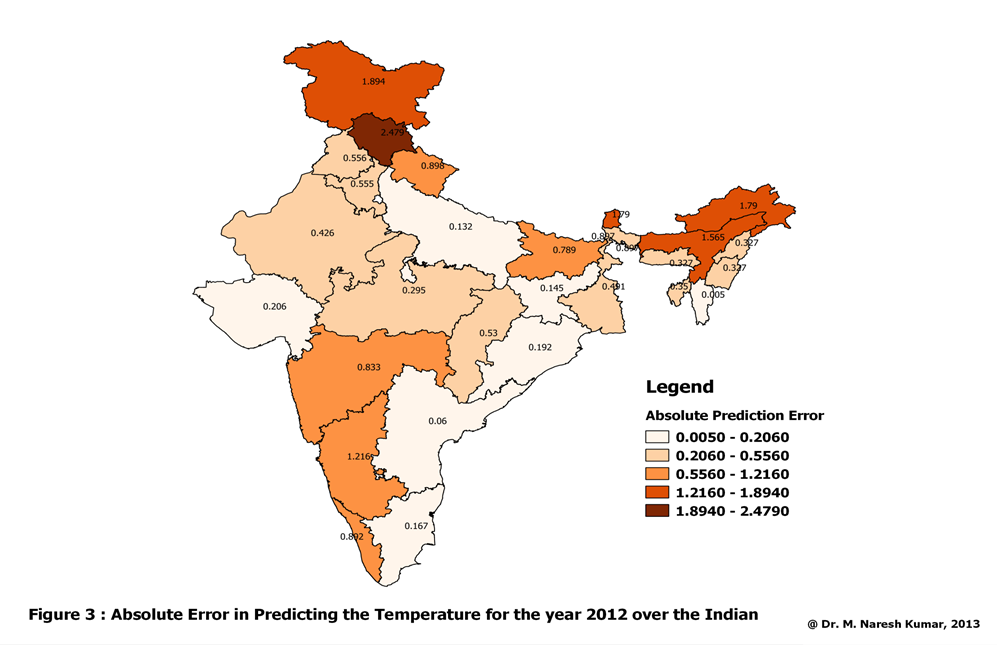}\\
  \caption{Absolute error in for predicting the temperature the year $2012$ over Indian region}\label{fig4}
\end{figure}

The spatial maps of the predicted temperature and the absolute error for the year $2012$ over Indian region is shown in Figures~\ref{fig3},\ref{fig4}.

The EM clustering could identify all the zones as per the Koppen classification over Indian region. The proposed strategy when employed for predicting temperature has resulted in an RMSE of $1.19$ in the Montane climate region and $0.89$ in the Humid Sub Tropical region as compared to $2.9$ and $0.95$ respectively predicted using k-means and linear regression method.

\section{Conclusions and Discussion}
\label{concl}
The expectation maximization clustering could identify the different climate zones as per the Koppen classification over Indian region. It is observed that the regions of Uttaranchal , Sikkim and Arunachal Pradesh have been identified as a separate group by EM different from  the Montane climate zone as per Koppen classification. This needs further investigations and introspection. 

EM clustering and SVM performed better than k-means and linear regressions only in Humid subtropical and Montane climate zones. It is observed the EM performance degrades as the dimensionality of the data set increases due to numerical precision problems. 

The fast growing volume of climate datasets and its high-dimensionality requires development of novel methods for preprocessing and information extraction. The focus of our future work would be on the development of techniques for big data climate data analytics.

\newpage

\bibliographystyle{elsarticle-num}

\end{document}